\documentclass{article}

     \usepackage[preprint,nonatbib]{neurips_2020}

\usepackage[utf8]{inputenc} 
\usepackage[T1]{fontenc}    
\usepackage{hyperref}       
\usepackage{url}            
\usepackage{booktabs}       
\usepackage{amsfonts}       
\usepackage{nicefrac}       
\usepackage{microtype}      
\usepackage{graphicx}
\usepackage{xcolor}
\usepackage{multicol}
\usepackage{amsmath,amssymb,mathtools}
\usepackage{subcaption}
\usepackage{siunitx}
\usepackage{multirow}
\usepackage{algorithm}
\usepackage[noend]{algpseudocode}

\DeclareMathOperator{\supp}{supp}

\newcommand{\aedit}[1]{#1}

\usepackage[normalem]{ulem}
\newcommand{\norm}[1]{\left\lVert#1\right\rVert}

\title{Reinforcement Learning with Combinatorial Actions: An Application to Vehicle Routing}

\author{%
  Arthur Delarue\\
  MIT Operations Research Center\\
  Cambridge, MA \\
  \texttt{adelarue@mit.edu} \\
  \And
  Ross Anderson \\
  Google Research \\
  Cambridge, MA \\
  \texttt{rander@google.com} \\
  \And
  Christian Tjandraatmadja \\
  Google Research \\
  Cambridge, MA \\
  \texttt{ctjandra@google.com} \\
}

\begin{document}

\maketitle

\begin{abstract}
  Value-function-based methods have long played an important role in reinforcement learning. However, finding the best next action given a value function of arbitrary complexity is nontrivial when the action space is too large for enumeration. We develop a framework for value-function-based deep reinforcement learning with a combinatorial action space, in which the action selection problem is explicitly formulated as a mixed-integer optimization problem. As a motivating example, we present an application of this framework to the capacitated vehicle routing problem (CVRP), a combinatorial optimization problem in which a set of locations must be covered by a single vehicle with limited capacity. On each instance, we model an action as the construction of a single route, and consider a deterministic policy which is improved through a simple policy iteration algorithm.  Our approach is competitive with other reinforcement learning methods and achieves an average gap of 1.7\% with state-of-the-art OR methods on standard library instances of medium size.
\end{abstract}

\section{Introduction}

Reinforcement learning (RL) is a powerful tool that has made significant progress on hard problems. For instance, in games like Atari \cite{mnih2013playing} or Go \cite{silver2017mastering}, RL algorithms have devised strategies that significantly surpass the performance of experts, even with little to no knowledge of problem structure. The success of reinforcement learning has carried over to other applications such as robotics \cite{OpenAI2020} and recommender systems \cite{gauci2018horizon}. As they encompass ever more application areas, RL algorithms need to be adjusted and expanded to account for new problem-specific features.

One area in which RL has yet to make a convincing breakthrough is combinatorial optimization. There has been significant effort in recent years to apply RL frameworks to NP-hard combinatorial optimization problems \cite{vinyals2015pointer, bello2016neural, khalil2017learning}, including the traveling salesman problem (TSP) or more general vehicle routing problem (VRP) (see \cite{bengio2018machine} for a recent survey). Solving these problems produces a practical impact for many industries, including transportation, finance, and energy. Yet in contrast to \cite{mnih2013playing, silver2017mastering}, RL methods have yet to match the performance of expert implementations in this domain, such as the state-of-the-art Concorde TSP solver \cite{applegate2006traveling, applegate2006concorde}.

A major difficulty of designing an RL framework for combinatorial optimization is formulating the action space. Value-based RL algorithms typically require an action space small enough to enumerate, while policy-based algorithms are designed with continuous action spaces in mind \cite{Silver2014}. These twin requirements severely limit the expressiveness of the action space, thus placing the entire weight of the problem on the machine learning model. For instance, in \cite{nazari2018reinforcement}, the selected action space models advancing the current vehicle one city at a time. RL algorithms for combinatorial optimization must therefore rely on complex architectures such as pointer networks \cite{vinyals2015pointer,bello2016neural,nazari2018reinforcement} or graph embeddings \cite{khalil2017learning}.

This paper presents a different approach, in which the combinatorial complexity is captured not only by the learning model, but also by the formulation of the underlying decision problem. We focus on the Capacitated Vehicle Routing Problem, a classical combinatorial problem from operations research \cite{toth2014vehicle}, where a \aedit{single} capacity-limited vehicle must be assigned \aedit{one or more} routes \aedit{to} satisfy customer demands while minimizing total travel distance. Despite its close relation to the TSP, the CVRP is a much more challenging problem and optimal approaches do not scale past hundreds of cities. Our approach is to formulate it as a sequential decision problem where the state space is the set of unvisited cities, and the action space consists of feasible routes. We estimate the value function of a state (i.e., its cost-to-go) using a small neural network. The action selection problem is then itself combinatorial, with the structure of a Prize Collecting Traveling Salesman Problem (PC-TSP) with a knapsack constraint and a nonlinear cost on unvisited cities. Crucially, the PC-TSP is a much easier combinatorial problem than the CVRP, allowing us to tractably exploit some of the combinatorial structure of the problem.

The contributions in this paper are threefold.
\begin{enumerate}
    \item We present a policy iteration algorithm for value-based reinforcement learning with combinatorial actions. At the \emph{policy improvement} step, we train a small neural network with ReLU activations to estimate the value function from each state. At the \emph{policy evaluation} step, we formulate the action selection problem from each state as a mixed-integer program, in which we combine the combinatorial structure of the action space with the neural architecture of the value function by adapting the branch-and-cut approach described in \cite{anderson2020strong}.
    \item We apply this technique to develop a reinforcement learning framework for combinatorial optimization problems in general and the Capacitated Vehicle Routing Problem in particular. Our approach significantly differs from existing reinforcement learning algorithms for vehicle routing problems, and allows us to obtain comparable results with much simpler neural architectures.
    \item We evaluate our approach against several baselines on random and standard library instances, achieving an average gap against the OR-Tools routing solver of 1.7\% on moderately-sized problems. While we do not yet match state-of-the-art operations research methods used to solve the CVRP, we compare favorably with heuristics using oracles of equal strength to our action selection oracle and to other RL approaches for solving CVRP \cite{nazari2018reinforcement}.
\end{enumerate}

\section{Background and related work}

\textbf{Combinatorial action spaces.} Discrete, high-dimensional action spaces are common in applications such as natural language processing \cite{He2016} and text-based games \cite{Zahavy2018}, but they pose a challenge for standard RL algorithms \cite{Dulac-Arnold2019}, chiefly because enumerating the action space when choosing the next action from a state becomes impossible. Recent remedies for this problem include selecting the best action from a random sample \cite{He2016}, approximating the discrete action space with a continuous one \cite{Dulac-Arnold2015,He2016a}, or training an additional machine learning model to wean out suboptimal actions \cite{Zahavy2018}. None of these approaches guarantees optimality of the selected action, and projection-based approaches in particular may not be applicable when the structure of the action space is more complex. In fact, even continuous action spaces can prove difficult to handle: for example, safety constraints can lead to gradient-based methods providing infeasible actions. Existing ways to deal with this issue include enhancing the neural network with a safety layer \cite{Dalal2018}, or modifying the policy improvement algorithm itself \cite{Chow2018}. More recently, \cite{ryu2019caql} propose a framework to explicitly formulate the action selection problem using optimization in continuous settings. We note that though combinatorial action spaces pose a challenge in deep reinforcement learning, they have been considered before in approximate dynamic programming settings with linear or convex learners \cite{powell2007approximate}. \aedit{In this paper, we formulate and solve the problem of selecting an optimal action from a combinatorial action space using mixed-integer optimization.}

\textbf{Combinatorial optimization and reinforcement learning.} In recent years, significant work has been invested in solving NP-hard combinatorial optimization problems using machine learning, notably by developing new architectures such as pointer networks \cite{vinyals2015pointer} and graph convolutional networks \cite{Joshi2019}. Leveraging these architectures, reinforcement learning approaches have been developed for the TSP \cite{bello2016neural,khalil2017learning} and some of its vehicle routing relatives \cite{Kool2019}, including the CVRP \cite{nazari2018reinforcement}. Crucially, the CVRP is significantly more challenging than the closely related TSP. While TSPs on tens of thousands of cities can be solved to optimality \cite{applegate2009certification}, CVRPs with more than a few hundred cities are very hard to solve exactly \cite{uchoa2017new}, often requiring cumbersome methods such as branch-and-cut-and-price, and motivating the search for alternative solution approaches. \aedit{This work adopts a hybrid approach, casting a hard combinatorial problem (CVRP) as a sequence of easier combinatorial problems (PC-TSP) in an approximate dynamic programming setting.}

\textbf{Optimizing over trained neural networks.} A major obstacle for solving the action selection problem with an exponential discrete action space is the difficulty of finding a global extremum for a nonlinear function (as a neural network typically is) over a nonconvex set. One possible solution is to restrict the class of neural networks used to guarantee convexity \cite{Amos2016}, but this approach also reduces the expressiveness of the value function. A more general approach is to formulate the problem as a mixed-integer program (MIP), which can be solved using general-purpose algorithms \cite{anderson2020strong}. Using an explicit optimization framework allows greater modeling flexibility and has been successfully applied in planning settings \cite{say2017nonlinear,Wu2019} as well as in RL problems with a continuous action space \cite{ryu2019caql}. Mixed-integer optimization has also proven useful in neural network verification \cite{cheng2017maximum,tjeng2018evaluating}. \aedit{In this paper, we use neural networks with ReLU activations, leveraging techniques developed in \cite{anderson2020strong} to obtain a strong formulation of the action selection problem.}

\section{Reinforcement learning model for CVRP}

\subsection{Problem formulation}

A CVRP instance is defined as a set of $n$ cities, indexed from $0$ to $n-1$. Each city $i>0$ is associated with a demand $d_i$; the distance from city $i$ to city $j$ is denoted by $\Delta_{ij}$. City $0$ is called the depot and houses a \aedit{single} vehicle with capacity $Q$ (or equivalently, a fleet of identical vehicles). Our goal is to produce routes that start and end at the depot such that \aedit{each non-depot city is} visited \aedit{exactly once}, the total demand $d_i$ in the cities \aedit{along one route does not exceed the vehicle capacity} $Q$, and the total distance \aedit{of all} routes is minimized. We assume the number of \aedit{routes we can serve} is unbounded, and note \aedit{that} the distance-minimizing solution does not necessarily minimize the number of vehicles used.

We formulate CVRP as a sequential decision problem, where a \emph{state} $s$ corresponds to a set of as-yet-unvisited cities, and an \emph{action} $a$ is a feasible route starting and ending at the depot and covering at least one city. We represent states using a binary encoding where 0 indicates a city has already been visited and 1 indicates it has not, i.e., $\mathcal{S}=\{0,1\}^{n}$ (though by convention, we never mark the depot as visited). The action space $\mathcal{A}$ corresponds to the set of all partial permutations of $n-1$ cities. Note that the sizes of both the state space and the action space are exponential in $n$, even if we only consider actions with the shortest route with respect to their cities.

The dynamics of the decision problem are modeled by a deterministic transition function $T:\mathcal{S}\times\mathcal{A}\to\mathcal{S}$, where $T(s,a)$ is the set of remaining unvisited cities in $s$ after serving route $a$, and a cost function $C:\mathcal{A}\to\mathbb{R}$, indicating the cost incurred by taking action $a$. Since cities cannot be visited twice, $T(s,a)$ is undefined if $a$ visits a city already marked visited in $s$. For clarity we define $\mathcal{A}(s)\subseteq\mathcal{A}$ as the set of feasible actions from state $s$. The unique terminal state $s^{\text{term}}$ corresponds to no remaining cities except the depot. Finding the best CVRP solution is equivalent to finding \aedit{a} least-cost path from the initial state $s^{\text{start}}$ (where all cities are unvisited) to $s^{\text{term}}$.

We consider a deterministic policy $\pi:\mathcal{S}\to\mathcal{A}$ specifying the action to be taken from state $s$, and we let $\Pi$ designate the set of all such policies. The value of a state $V^{\pi}(s)$ is the cost incurred by applying $\pi$ repeatedly starting from state $s$, i.e., $V^{\pi}(s)=\sum_{t=1}^T C(a^t|a^t=\pi(s^{t-1}), s^t=T(s^{t-1}, a^t), s^0=s, s^T=s^{\text{term}})$. An optimal policy $\pi^*$ satisfies $\pi^*=\arg\min_{\pi\in\Pi}V^{\pi}(s^{\text{start}})$.

Given a starter policy $\pi_0$, we repeatedly improve it using a simple \emph{policy iteration} scheme. In the $k$-th \emph{policy evaluation} step, we repeatedly apply the current policy $\pi_{k-1}(\cdot)$ from $N$ randomly selected start states $\{s^{0,i}\}_{i=1}^N$. For each random start state $s_{0,i}$, we obtain a \emph{sample path}, i.e. a finite sequence of action-state pairs $(a^{1,i},s^{1,i}), \ldots, (a^{T,i}, s^{T,i})$ such that $s^{t,i}=T(s^{t-1,i}, a^{t,i})$, and $s^{T,i}=s^{\text{term}}$, as well as the cumulative cost $c^{t,i}=\sum_{t'=t}^TC(a^{t',i})$ incurred from each state in the sample path. In the \emph{policy improvement} step, we use this data to train a small neural network $\hat{V}^{\aedit{k-1}}(\cdot)$ to approximate the value function $V^{\pi_{k-1}}(\cdot)$. This yields a new policy $\pi_k(\cdot)$ defined using the Bellman rule:
\begin{equation}
    \pi_k(s)=\arg\min_{a\in\mathcal{A}\aedit{(s)}}C(a) + \hat{V}^{\aedit{k-1}}(t\coloneqq T(s, a)).
    \label{eq:bellman}
\end{equation}

\subsection{Value function learning}

The approximate policy iteration approach described above is both on-policy and model-based:\footnote{\aedit{In fact, the state transition model is deterministic and we have perfect knowledge of it.}} in each iteration, our goal is to find a good approximation for the value function $V^{\pi}(\cdot)$ of the current policy $\pi(\cdot)$. The value function $V^{\pi}(s)$ of a state $s$ is the cost of visiting all remaining cities in $s$, i.e., the cost of solving a smaller CVRP instance over the unvisited cities in $s$ using policy $\pi(\cdot)$.

In our approximate dynamic programming approach, the value function captures much of the combinatorial difficulty of the vehicle routing problem, so we model $\hat{V}$ as a small neural network with a fully-connected hidden layer and rectified linear unit (ReLU) activations. We train this neural network to minimize the mean-squared error (MSE) on the cumulative cost data gathered at the policy evaluation step. To limit overfitting, we randomly remove some data (between 10\% and 20\%) from the training set and use it to evaluate the out-of-sample MSE.

We select initial states by taking a single random action from $s^{\text{start}}$, obtained by selecting the next city uniformly at random (without replacement) until we select a city that we cannot add without exceeding vehicle capacity. If this procedure does not allow for enough exploration, we may gather data on too few states and overfit the value function. We therefore adapt a technique from approximate policy iteration \cite{Lagoudakis2003,Scherrer2014}, in which we retain data from one iteration to the next, and exponentially decay the importance of data gathered in previous iterations \cite{bertsekas1996neuro}. Calling $(s^{(i,k')}, c^{(i, k')})$ the $i$-th data point (out of $N_{k'}$) from iteration $k'$, the training objective in iteration $k$ becomes $\sum_{k'=0}^{k}\sum_{i=1}^{N_{k'}}\gamma^{k-k'}(\hat{V}(s^{(i,k')}) - c^{(i, k')})^2$, where $\gamma\in(0,1]$ denotes the \emph{retention factor}.

\subsection{Policy evaluation with mixed-integer optimization}

The key innovation in our approach comes at the policy evaluation step, where we repeatedly apply policy $\pi_{k}(\cdot)$ to random start states until we reach the terminal state. Applying policy $\pi_{k}(\cdot)$ to state $s$ involves solving the optimization problem in Eq.~\eqref{eq:bellman}, corresponding to finding a capacity-feasible route minimizing the sum of the immediate cost (length of route) and the cost-to-go (value function of new state). This is a combinatorial optimization problem (PC-TSP) that cannot be solved through enumeration but that we can model as a mixed-integer program. Without loss of generality we consider problem~\eqref{eq:bellman} for a state $s$ such that $s_i=1$ for $1 \le i \le m$, and $s_i=0$ for $m < i < n$, i.e. such that only the first $m$ cities remain unvisited. We then obtain the following equivalent formulation for problem~\eqref{eq:bellman}:
\begin{subequations}
\label{eq:formulation}
\begin{align}
    \min\quad & \sum_{i=0}^{m}\sum_{j=0;j\neq i}^{m} \Delta_{ij}x_{ij} + \hat{V}(t)\label{eq:objective}\\
    \text{s.t.}\quad & \sum_{j=0;j\neq i}^{m}x_{ij}=y_i & \forall\; 0\le i \le m\label{eq:flow-out}\\
    & \sum_{j=0;j\neq i}^{m}x_{ji}=y_i & \forall\; 0\le i \le m\label{eq:flow-in}\\
    & \sum_{i=1}^{m}d_iy_i\le Q\label{eq:capacity}\\
    & \sum_{i\in S}\sum_{j\in \{0, \ldots, m\}\backslash S}x_{ij}\ge y_i&\forall\; S\subseteq\{1,\ldots,m\}, i\in S\label{eq:cutset-out}\\
    & \sum_{i\in S}\sum_{j\in \{0, \ldots, m\}\backslash S}x_{ji}\ge y_i&\forall\; S\subseteq\{1,\ldots,m\}, i\in S\label{eq:cutset-in}\\
\end{align}
\begin{align}
    & t_i = \begin{cases}1-y_i,& 0 \le i \le m\\0,& m < i < n\end{cases} \label{eq:auxiliary}\\
    & x_{ij}\in\{0,1\}&\forall\; 0 \le i \neq j \le m\\
    & y_0=1\\
    & y_i\in\{0,1\}&\forall\; 1 \le i \le m.
\end{align}
\end{subequations}
The binary decision variable $y_i$ is 1 if city $i$ is included in the route, and 0 otherwise, and the binary decision variable $x_{ij}$ is 1 if city $i$ immediately precedes city $j$ in the route (we constrain $y_0$ to be 1 because the depot must be included in any route). The binary decision variable $t_i$ is 1 if city $i$ remains in the new state $T(s,a)$ and 0 otherwise. \aedit{The objective~\eqref{eq:objective} contains the same two terms as~\eqref{eq:bellman}, namely the total distance of the next route, plus the future cost associated with leaving some cities unvisited.}

Constraint~\eqref{eq:flow-out} (resp. \eqref{eq:flow-in}) ensures that if city $i$ is included in the route, it must be immediately followed (resp. preceded) by exactly one other city (flow conservation constraints). Constraint~\eqref{eq:capacity} imposes that the total demand among selected cities does not exceed the vehicle's capacity. Constraints~\eqref{eq:cutset-out} and \eqref{eq:cutset-in} ensure that the route consists of a single tour starting and ending at the depot; they are called \emph{cycle-breaking} or \emph{cutset} constraints and are a standard feature of MIP formulations of routing problems. Finally, constraints~\eqref{eq:auxiliary} relate the action selection variable $y_i$ for each city $i$ to the corresponding new state variables $t_i$. 

We make two further comments regarding problem~\eqref{eq:formulation}. First, we note that the formulation includes an exponential number of constraints of type~\eqref{eq:cutset-out} and \eqref{eq:cutset-in}. This issue is typically addressed via ``lazy constraints'', a standard feature of mixed-integer programming solvers such as Gurobi or SCIP, in which constraints are generated as needed. In practice, the problem can be solved to optimality without adding many such constraints. Second, as currently formulated, problem~\eqref{eq:formulation} is not directly a mixed-integer linear program, because $\hat{V}(t)$ is a nonlinear function. However, because we choose ReLU-activated hidden layers, it turns out it is a piecewise linear function which can be represented using additional integer and continuous variables and linear constraints \cite{say2017nonlinear,anderson2020strong,ryu2019caql}.

In the worst case, problem~\eqref{eq:formulation} can be solved in exponential time in $m$. But modern MIP solvers such as Gurobi and SCIP use techniques including branch-and-bound, primal and dual heuristics, and cutting planes to produce high-quality solutions in reasonable time \cite{scip,gurobi}. Given enough time, MIP solvers will not just return a solution but also certify optimality of the selected action.

A MIP approach for action selection was presented in \cite{ryu2019caql}, but ours differs in several key ways. First, we consider a discrete action space instead of a continuous one. Second, we estimate the cost-to-go directly from cumulative costs rather than using a Q-learning approach because (i) state transitions in our problem are linear in our action and (ii) \aedit{computing cumulative costs allows us to avoid solving the combinatorial action selection problem at training time, reducing overall computation. Separating training and evaluation may yield additional computational benefits, since} the training loop \aedit{could benefit from} a gradient-descent-friendly architecture and hardware such as a GPU/TPU, while the evaluation loop relies on CPU-bound optimization solvers. Separating the tasks allows for better parallelization and hardware specification.

To further take advantage of the combinatorial structure of the problem, we augment the objective~\eqref{eq:objective} with known lower bounds. The cost-to-go of a state cannot be lower than the distance required to serve the farthest remaining city from the depot, nor can it be exceeded by the summed lengths of each remaining city's shortest incoming edge. Given such lower bounds $\{LB^p(t)\}_{p=1}^P$, we replace $\hat{V}(t)$ in the objective with $\max(\hat{V}, LB^1(t), \ldots, LB^P(t))$. As long as the lower bounds are linear in $t$ (as are the ones we mention), the addition does not significantly increase solve time for problem~\eqref{eq:formulation}.

\section{Computational results}

We now present results on benchmark instances from the operations research literature and on random instances from \cite{nazari2018reinforcement}. We compare our results to several other approaches, and analyze their sensitivity to input parameters. Our implementation is in C++ and we use SCIP 6.0.2 as a MIP solver.

\subsection{Comparison with existing methods \aedit{and runtime analysis}}

In order to compare the results of our approach to existing RL algorithms, we consider random instances described in \cite{nazari2018reinforcement}, in which cities are sampled uniformly at random from the unit square, distances are Euclidean, and demands are sampled uniformly at random from $\{1,\ldots,9\}$. We present results on instances with 11, 21 and 51 cities in Table~\ref{tab:comparison}.

We note that our solutions compare favorably with both simple heuristics and existing RL approaches, and nearly match the performance of state-of-the-art operations research methods. We must, however, qualify the comparison with a caveat: existing RL approaches for CVRP \cite{nazari2018reinforcement, Kool2019} consider a distributional setting where learning is performed over many instances and evaluated out of sample, whereas we consider a single-instance setting. Though these approaches are different, they nevertheless provide a useful comparison point for our method.

\begin{figure}
    \centering
    \includegraphics[width = 0.45\columnwidth]{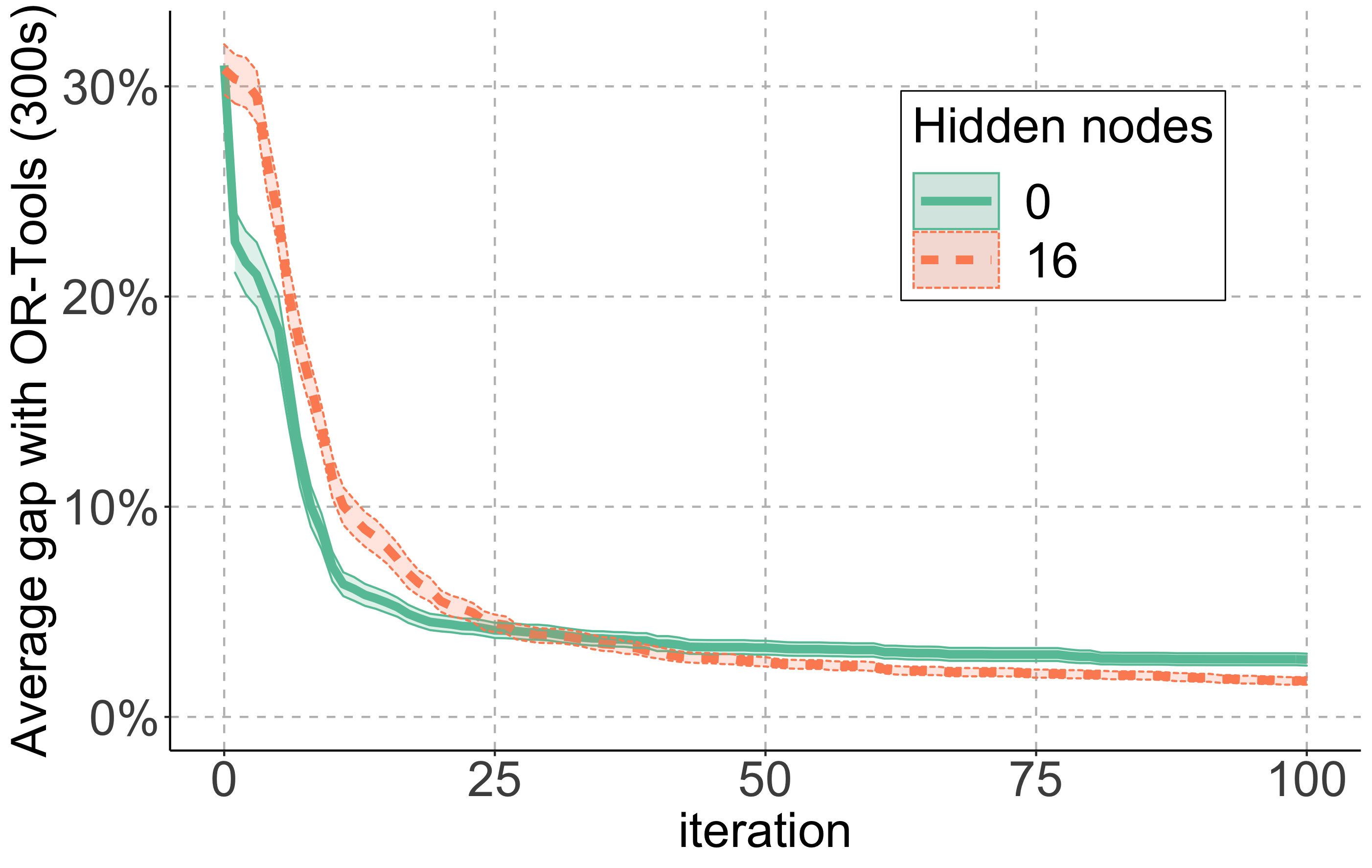}
    \caption{Results of our approach on 50 instances from the CVRP library (A and B) \cite{uchoa2017new}, ranging in size from 32 to 78 cities. We perform 100 policy iterations on each instance.}
    \label{fig:cvrplib}
\end{figure}

One advantage of our method is that it can be used on instances without distributional information, which may be closer to a real-world setting. In Figure~\ref{fig:cvrplib}, we evaluate our performance on standard library instances from CVRPLIB \cite{uchoa2017new,augerat1995computational}, comparing our results with those of OR-Tools. \aedit{Over 50 instances, the average final gap against OR-Tools is just 1.7\%.}

\begin{table}
    \centering
    \caption{Comparison of results with existing approaches: a ``greedy'' approach in which each action is selected to minimize immediate distance traveled plus a trivial upper bound on the cost-to-go; results from Nazari et al. \cite{nazari2018reinforcement} and Kool et al. \cite{Kool2019}; our own approach (RLCA) with 16 neurons; OR-Tools routing solver with 60s of guided local search (300s for $n=51$)\aedit{; and an optimal approach using the OR-Tools CP-SAT constraint programming solver (does not solve to optimality within 6 hours for $n = 51$). We report the mean total distance ($\mu$) and standard error on the mean $\sigma_{\mu}$, over 1000 random instances for each value of $n$.}}
    \label{tab:comparison}
    \begin{tabular}{lcccccc}
        \toprule
        & \multicolumn{2}{c}{$n=11$} & \multicolumn{2}{c}{$n=21$} & \multicolumn{2}{c}{$n=51$}\\
        Method & $\mu$ & $\sigma_{\mu}$ & $\mu$ & $\sigma_{\mu}$ & $\mu$ & $\sigma_{\mu}$\\
        \midrule
        Greedy & 4.90 & 0.03 & 7.16 & 0.03 & 13.55 & 0.04\\
        Nazari et al. \cite{nazari2018reinforcement}& 4.68 & 0.03 & 6.40 & 0.03 & 11.15 & 0.04\\
        Kool et al. \cite{Kool2019}& - & - & 6.25 & - & 10.62 & - \\
        RLCA-16 & 4.55 & 0.03 & 6.16 & 0.03 & 10.65 & 0.04\\
        OR-Tools \cite{or-tools} & 4.55 & 0.03 & 6.13 & 0.03 & 10.47 & 0.04\\
        Optimal & 4.55 & 0.03 & 6.13 & 0.03 & - & -\\
         \bottomrule
    \end{tabular}
\end{table}

\begin{table}
    \caption{Average time in seconds to solve a single action selection problem, with the MIP solvers Gurobi 8.1 and SCIP 6.0.2, averaged over 50 random instances. Instances are from the first policy iteration after taking a single random action, trained using 5000 path evaluations.}
    \label{tab:runtime}
        \hfill
        \begin{subtable}{0.45\columnwidth}
        \caption{$n = 21$}
        \centering
        \begin{tabular}{ccc}
            \toprule
            Hidden & \multicolumn{2}{c}{Runtime (s)} \\
            nodes & SCIP & Gurobi\\
            \midrule
            0 (LR) & 0.059 & 0.0085 \\
            4 & 0.84 & 0.16\\
            16 & 3.1 & 0.38\\            \bottomrule
        \end{tabular}
        \end{subtable}\hfill%
        \begin{subtable}{0.45\columnwidth}
        \caption{$n = 51$}
        \centering
        \begin{tabular}{ccc}
            \toprule
            Hidden & \multicolumn{2}{c}{Runtime (s)} \\
            nodes & SCIP & Gurobi\\
            \midrule
            0 (LR) & 3.19 & 0.097\\
            4 & 132.5 & 11.9\\
            16 & 234.7 & 39.3\\
            \bottomrule
        \end{tabular}
        \end{subtable}\hfill
\end{table}

\aedit{It is of interest to study our method's runtime. At training time, our two main operations are evaluating many sample paths with the current policy, and re-training the value function estimator (at evaluation time, we just evaluate one sample path). With up to 16 hidden ReLUs, neural net training is fast, so the bottleneck is solving \eqref{eq:formulation}. We cannot provide precise runtimes because our experiments were executed in parallel on a large cluster of machines (which may suffer from high variability), but we present average MIP runtimes given different architectures and solvers in Table~\ref{tab:runtime} and, based on these values, we describe how to calculate an estimate of the runtimes. The runtime of a policy iteration is $(\#\text{ sample paths})\times(\#\text{ MIPs per path})\times (\text{MIP runtime})$. For $n=21$ cities (16 hidden nodes), SCIP solves the average MIP in $\sim3$s (Gurobi in $\sim0.4$s), and we almost never exceed 10 MIPs per path, so computing a policy iteration with 250 sample paths takes about 2h using SCIP (15min using Gurobi). We can reduce runtime with parallelism: with as many machines as sample paths, the SCIP running time becomes about 30s (plus parallel pipeline overhead). For $n=51$, SCIP is slower ($\sim 240$s per MIP), and a policy iteration may take up to an hour in parallel. In contrast, Nazari et al.'s \cite{nazari2018reinforcement} runtime bottleneck is neural net training (hours), but evaluation is much faster (seconds).}

\subsection{Ablation studies and sensitivity analysis}

As a success metric, we evaluate the quality of a solution $x$ using the gap $g$ between $x$ and the \aedit{provably optimal solution $x_{\text{CP-SAT}}$ obtained by CP-SAT} on the same instance, where $g=(x - x_{\text{CP-SAT}})/x_{\text{CP-SAT}}$.

\begin{figure}
    \centering
    \begin{subfigure}{0.45\columnwidth}
        \includegraphics[width=\columnwidth]{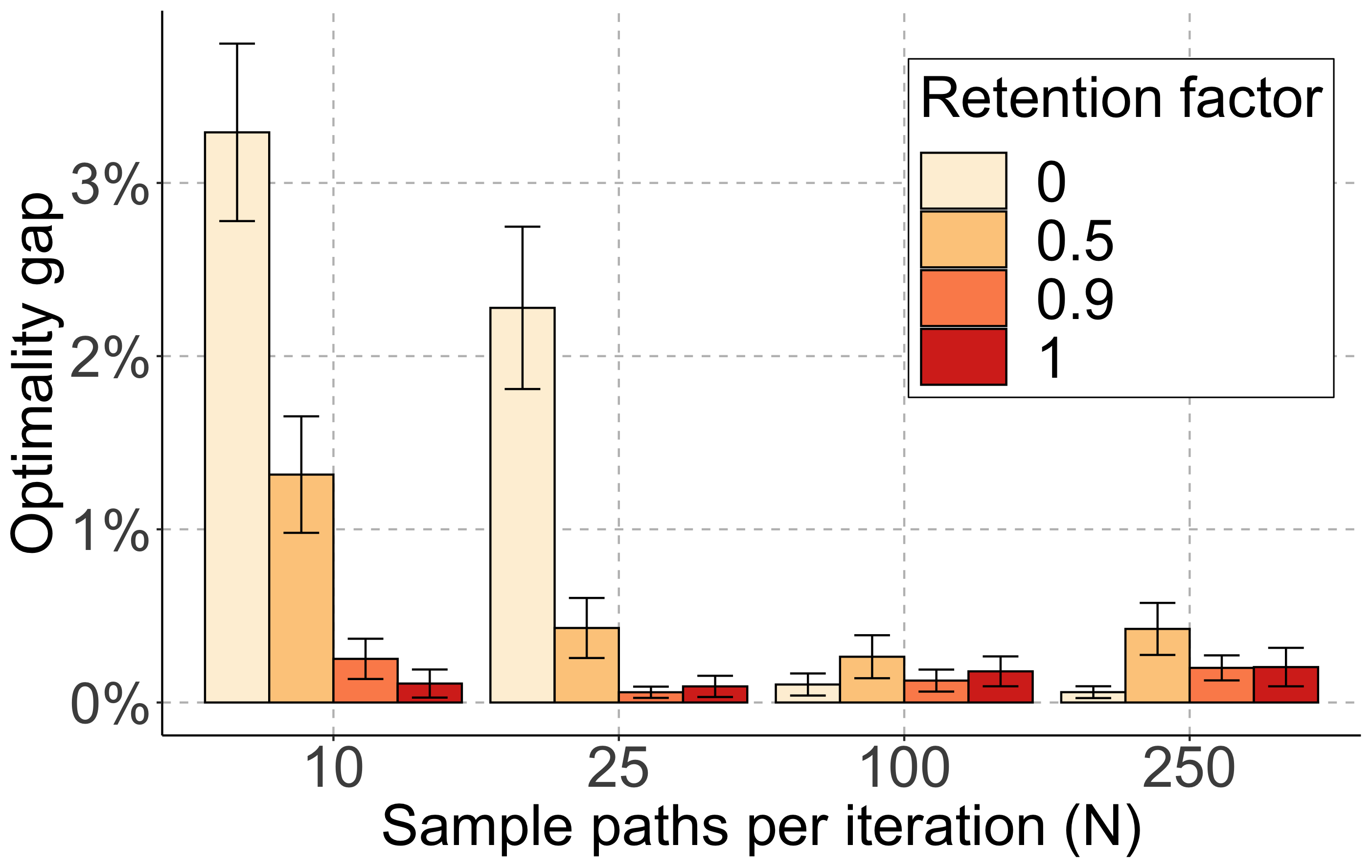}
        \caption{Gap for varying amounts of data and retention factors after 30 policy iterations ($n = 11$).}
        \label{fig:num_evals_11}
    \end{subfigure}\hfill
    \begin{subfigure}{0.45\columnwidth}
        \includegraphics[width=\columnwidth]{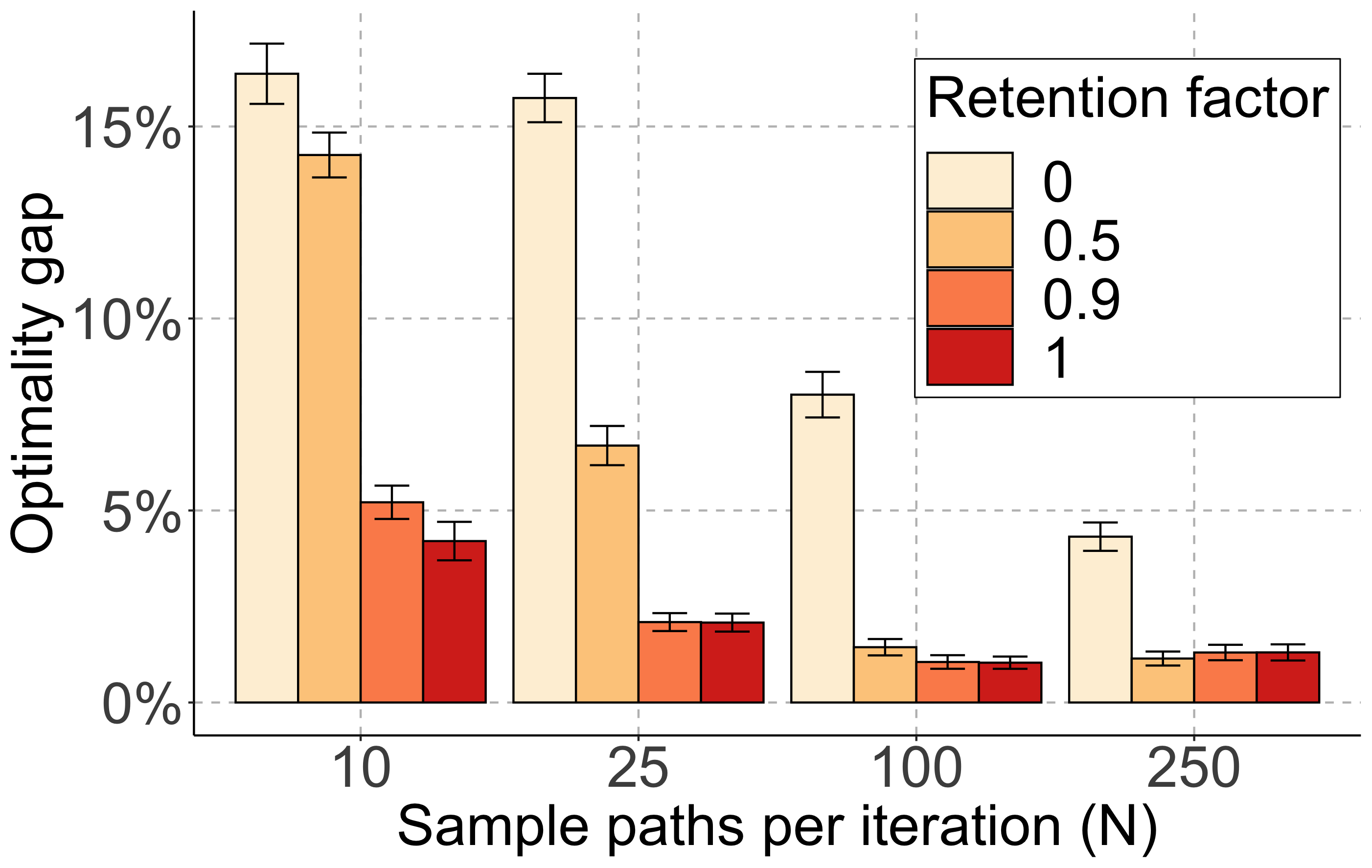}
        \caption{Gap for varying amounts of data and retention factors after 30 policy iterations ($n = 21$).}
        \label{fig:num_evals_21}
    \end{subfigure}\hfill
    \begin{subfigure}{0.45\columnwidth}
        \includegraphics[width=\columnwidth]{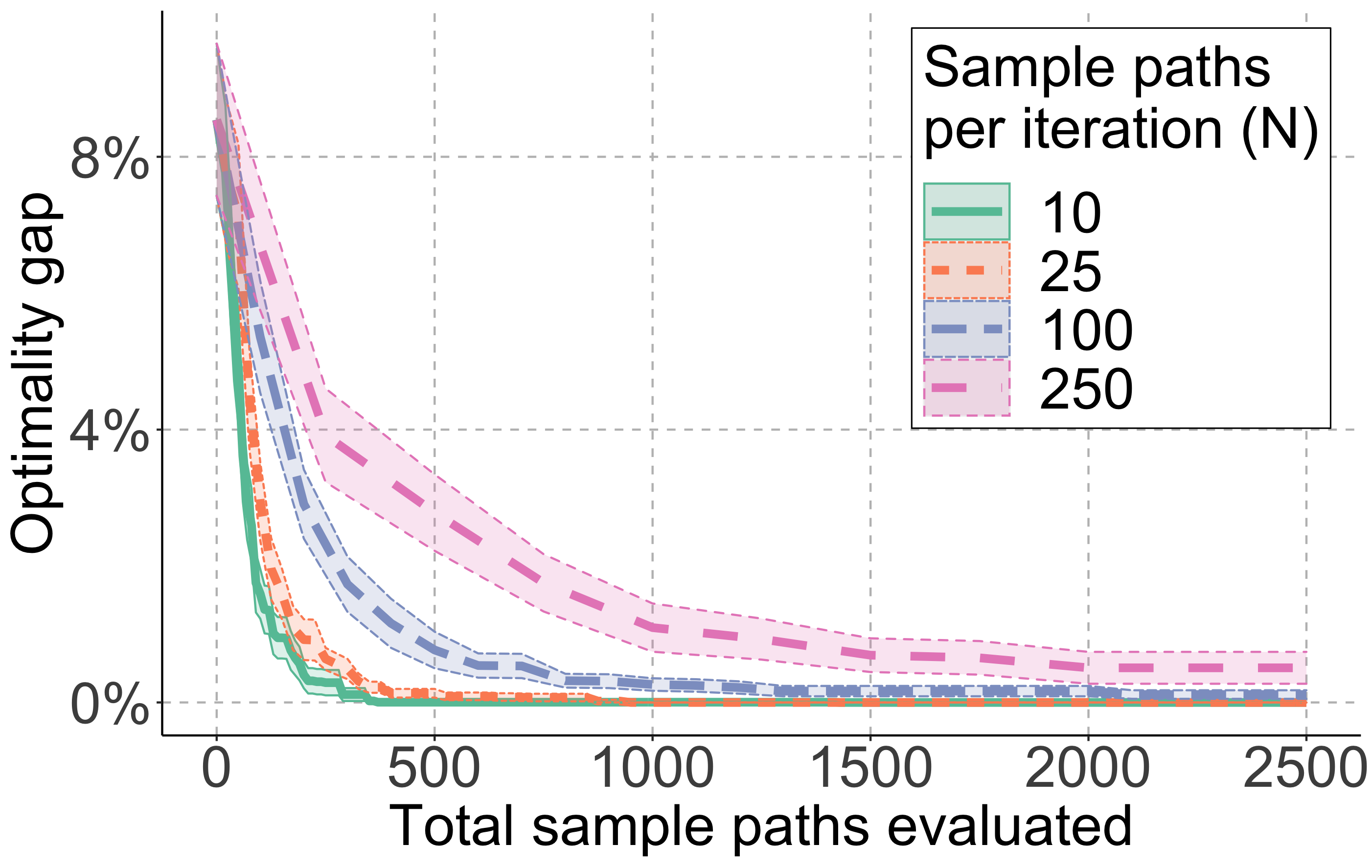}
        \caption{Gap as a function of total evaluated sample paths, with retention factor $\gamma = 1$ ($n = 11$).}
        \label{fig:cumulative_evals_11}
    \end{subfigure}\hfill
    \begin{subfigure}{0.45\columnwidth}
        \includegraphics[width=\columnwidth]{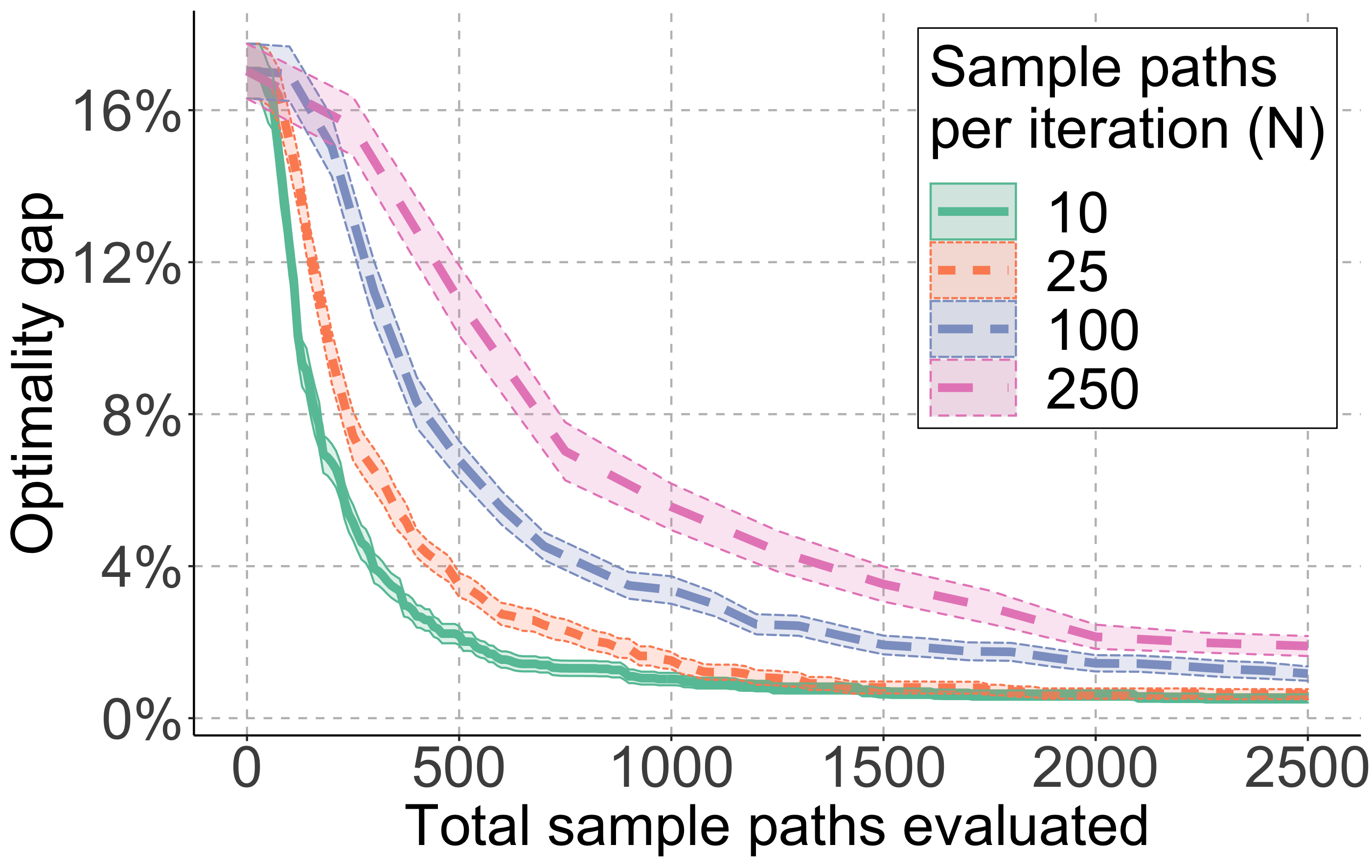}
        \caption{Gap as a function of total evaluated sample paths, with retention factor $\gamma=1$ ($n = 21$).}
        \label{fig:cumulative_evals_21}
    \end{subfigure}
    \caption{Effect of the amount of data on performance. Results averaged over 50 random Euclidean instances with 11 or 21 cities, error bars indicate standard errors (SEM).}
    \label{fig:data_requirement}
\end{figure}

\paragraph{Data requirements.} At each policy evaluation step, we evaluate the current policy from $N$ randomly selected start states to obtain sample paths. By evaluating more sample paths, we can use more data to update the policy, but we also require more computing resources. In Figs.~\ref{fig:num_evals_11} and \ref{fig:num_evals_21}, we show the solution quality obtained after 30 policy iterations, as we vary the number of sample paths evaluated per iteration $N$ and the retention factor $\gamma$. For a fixed number of iterations, results improve with more data, with diminishing returns. Keeping data between policy iterations significantly improves the solution quality, especially with few evaluations per iteration and little to no decay.

Even with just 25 evaluations per iteration, keeping data from one iteration to the next leads to solutions with a very small gap. In Figs.~\ref{fig:cumulative_evals_11} and \ref{fig:cumulative_evals_21}, we show the solution quality when keeping data between iterations (with $\gamma=1$) as a function of the cumulative number of evaluations and the number of evaluations per iteration. We obtain the best results with 250 iterations and 10 evaluations per iteration, suggesting that rapid iterations may be preferable on small instances.

\begin{figure}
    \centering
    \begin{subfigure}{0.45\columnwidth}
        \includegraphics[width=\columnwidth]{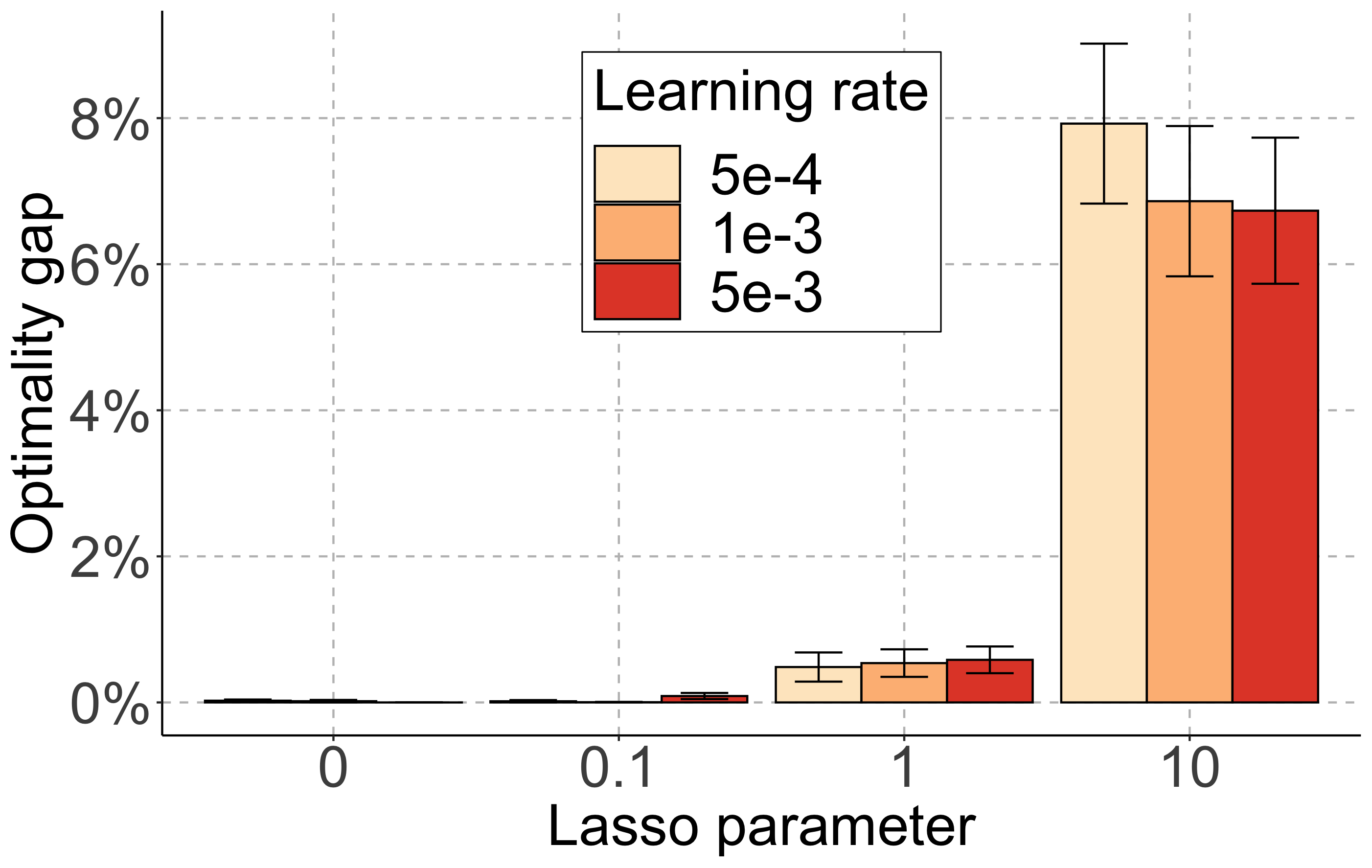}
        \caption{Gap after 50 policy iterations (n = 11).}
        \label{fig:l1_gap_11}
    \end{subfigure}\hfill
    \begin{subfigure}{0.45\columnwidth}
        \includegraphics[width=\columnwidth]{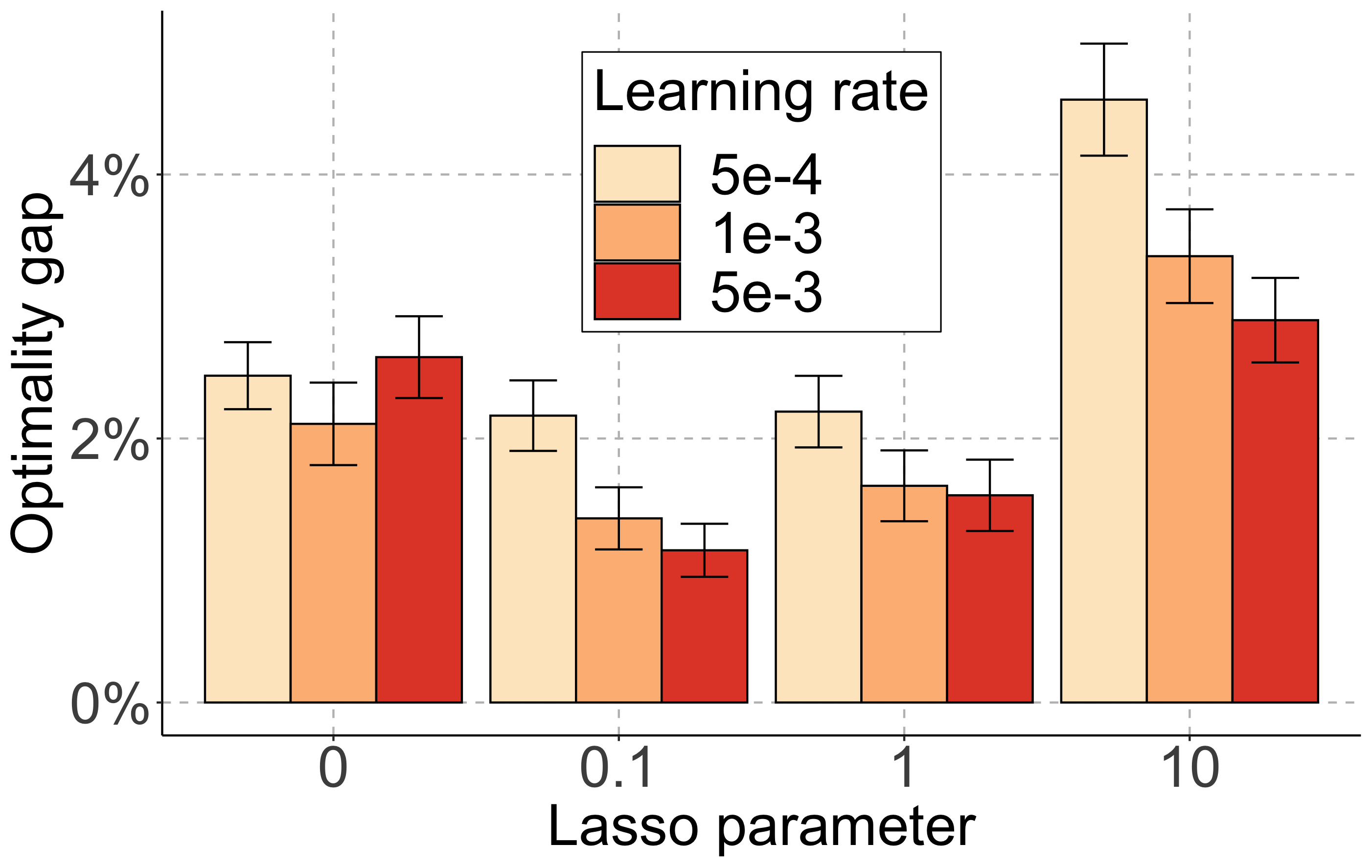}
        \caption{Gap after 50 policy iterations (n = 21).}
        \label{fig:l1_gap_21}
    \end{subfigure}\hfill
    \caption{Effect of training parameters (lasso weight and learning rate) on performance. Results averaged over 50 random Euclidean instances (11 or 21 cities), error bars indicate standard errors (SEM).}
    \label{fig:training}
\end{figure}

\begin{figure}
    \centering
    \begin{subfigure}{0.45\columnwidth}
        \includegraphics[width=\columnwidth]{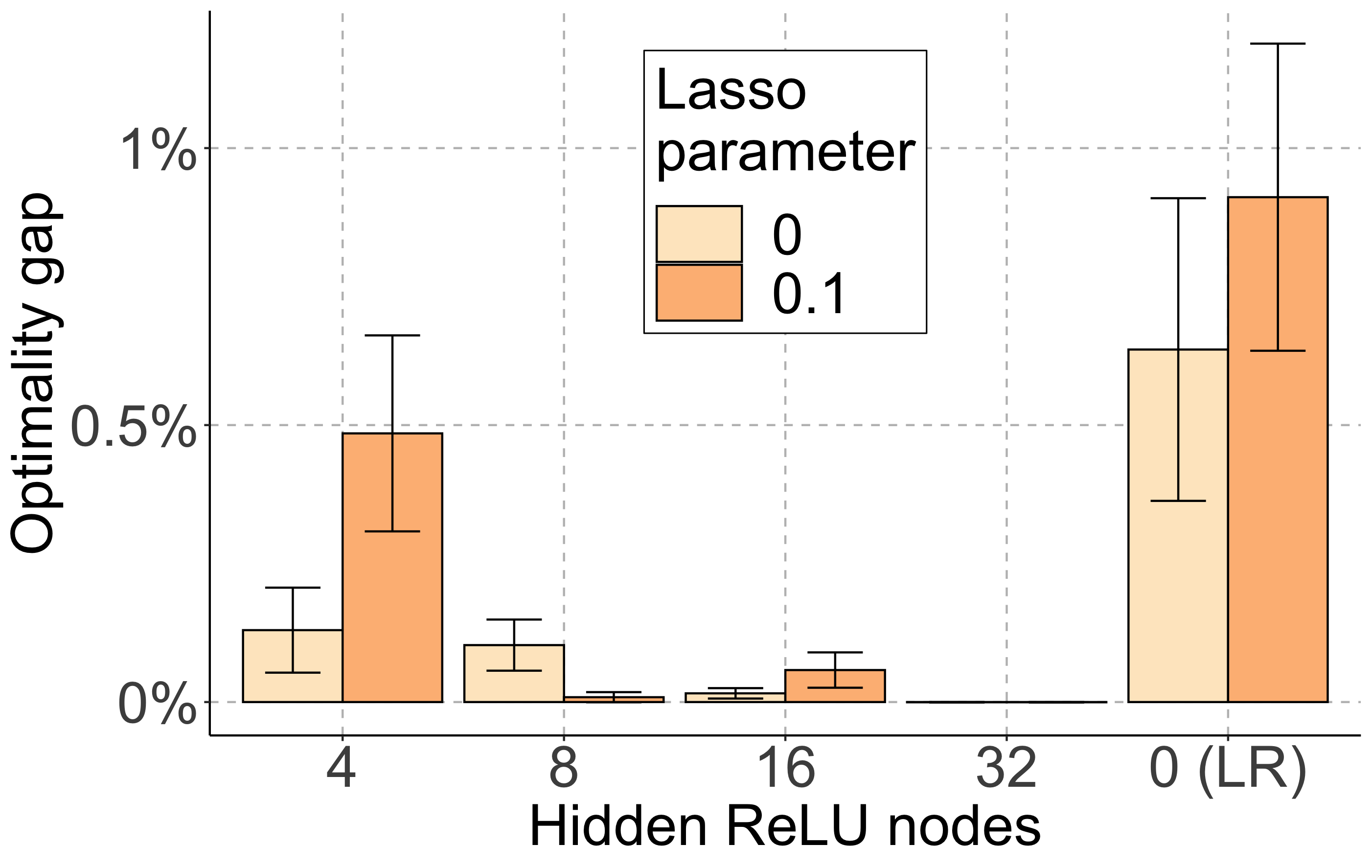}
        \caption{Gap after 50 policy iterations (n = 11).}
        \label{fig:arch_11}
    \end{subfigure}\hfill
    \begin{subfigure}{0.45\columnwidth}
        \includegraphics[width=\columnwidth]{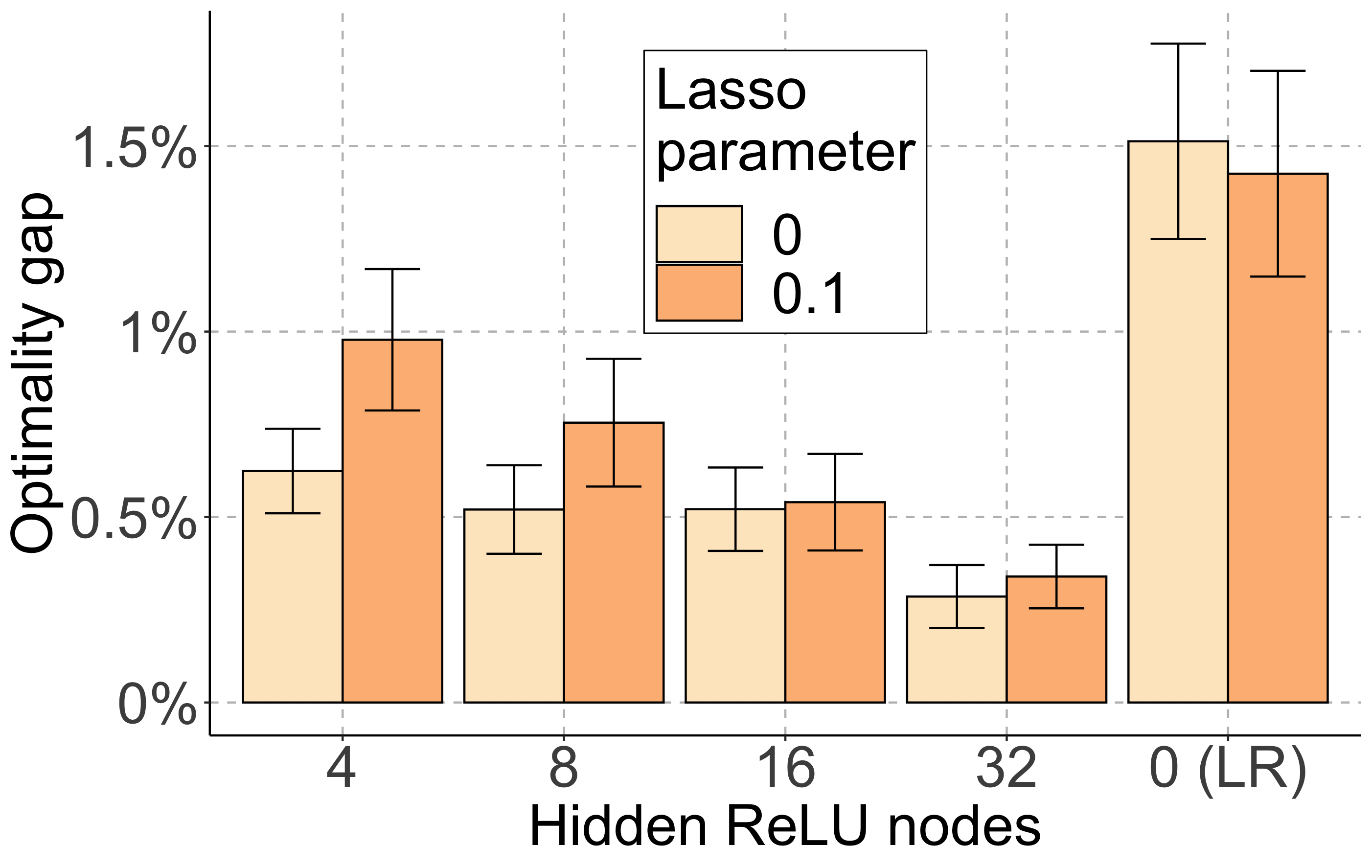}
        \caption{Gap after 250 policy iterations (n = 21).}
        \label{fig:arch_21}
    \end{subfigure}\hfill
    \caption{Effect of neural network size on performance (0 hidden nodes corresponds to linear regression). Results averaged over 50 random Euclidean instances, error bars show standard errors (SEM).}
    \label{fig:architecture}
\end{figure}

\paragraph{Architecture, training and regularization.} In Figure~\ref{fig:training}, we consider a neural network with 16 hidden ReLU nodes and show the effect of the batch SGD learning rate and the LASSO regularization parameter $\lambda$ (given all neural network weights as a vector $w$ of length $|w|$, we penalize the training objective with the LASSO term $(\lambda/|w|) \norm{w}_1$). We notice that mild regularization does not adversely impact performance: this is particularly important in our setting, because a sparser neural network can make the action selection problem MIP formulation easier to solve. In Figure~\ref{fig:architecture}, we analyze the effects of increasing neural network size on the solver performance. We note that even 4 neurons leads to a significant improvement over a linear model; on small instances, a 32-neuron network leads to optimal performance.

\begin{figure}
    \centering
    \begin{subfigure}{0.45\columnwidth}
        \includegraphics[width=\columnwidth]{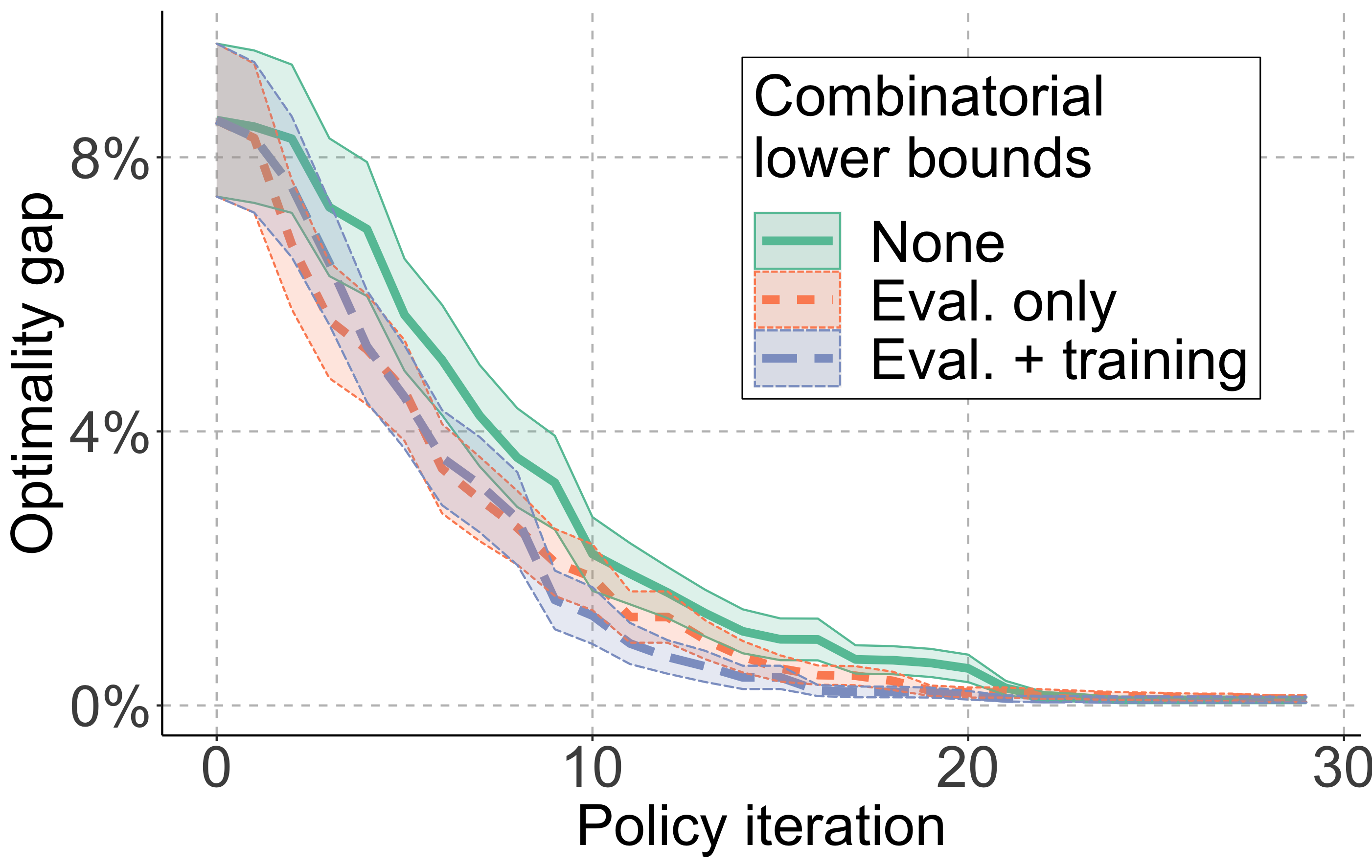}
        \caption{$n=11$}
    \end{subfigure}\hfill
    \begin{subfigure}{0.45\columnwidth}
        \includegraphics[width=\columnwidth]{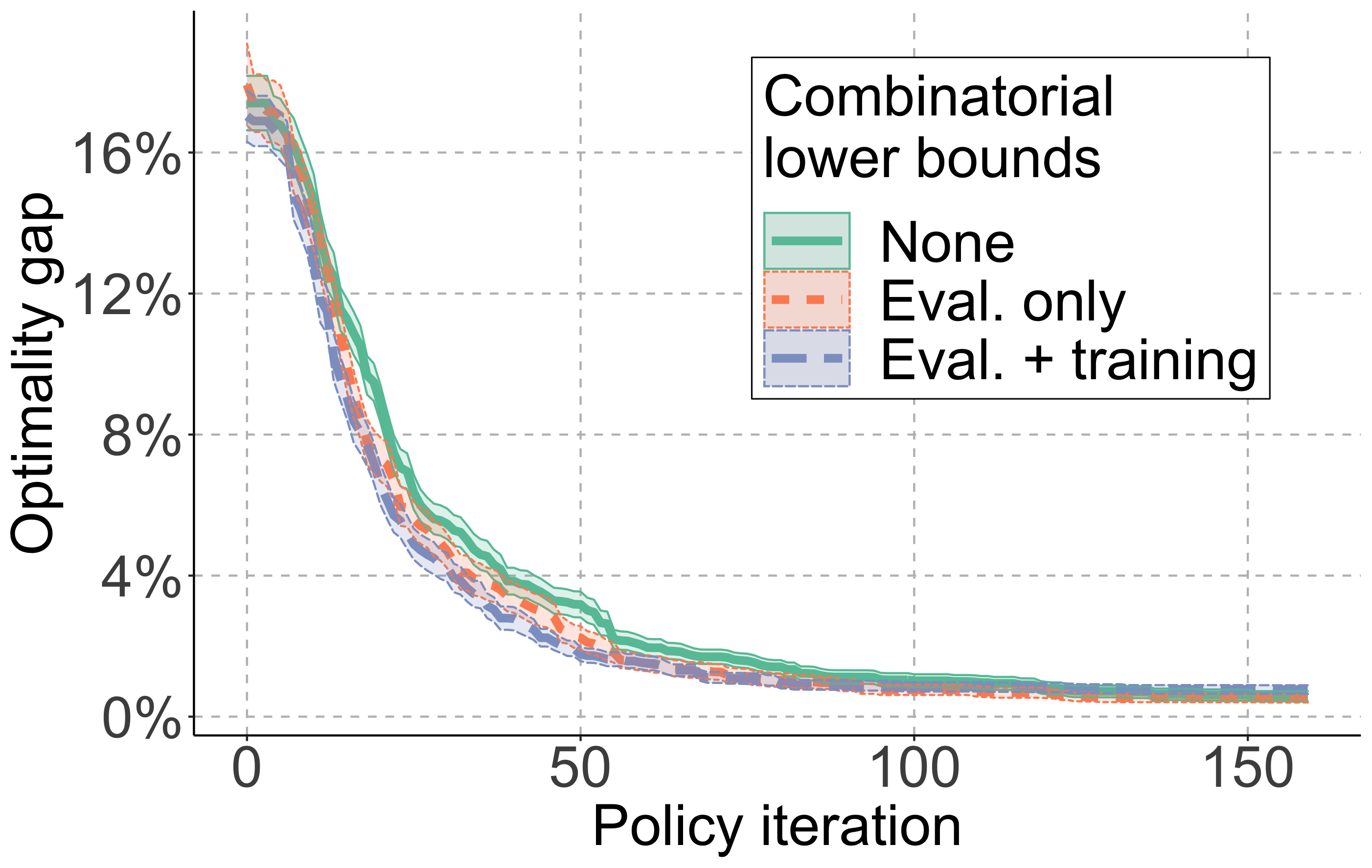}
        \caption{$n=21$}
    \end{subfigure}
    \caption{Effect of including lower bounds in our RL framework. Results averaged over 50 instances.}
    \label{fig:comblb}
\end{figure}

\paragraph{Combinatorial lower bounds.} As mentioned above, we augment the neural network modeling the value function with known combinatorial lower bounds. We can include them in the action selection problem, but also during training by replacing $\hat{V}(s)$ with $\max(\hat{V}(s), LB^1(s), \ldots, LB^P(s))$ in the training objective. Figure~\ref{fig:comblb} shows that including these lower bounds provide a small but nonetheless significant improvement to the convergence of our policy iteration scheme.

\section{Conclusion and Discussion}

We have presented a novel RL framework for the Capacitated Vehicle Routing Problem (CVRP). Our approach is competitive with existing RL approaches, and even with problem-specific approaches on small instances. A combinatorial action space allows us to leverage the structure of the problem to develop a method that combines the best of reinforcement learning and operations research.

A big difference between this method and other RL approaches \cite{nazari2018reinforcement,Kool2019} is that we consider a single instance at a time, instead of learning a single model for multiple instances drawn from a similar distribution. Single-instance learning is useful because the distribution of real-world problems is often unknown and can be hard to estimate because of small sample sizes. However, a natural extension of this work is to consider the multi-instance problem, in which learning is performed offline, and a new instance can be solved via a single sequence of PC-TSPs.

\aedit{Another possible extension of this work is the consideration of other combinatorial optimization problems. A simple next step would be to impose an upper bound on the number of routes allowed, by augmenting the state space to keep track of the number of past routes and penalizing states where this number is too high. Beyond this simple example lies a rich landscape of increasingly complex vehicle routing problems \cite{toth2014vehicle}. The success of local search methods in tackling these problems suggests an orthogonal reinforcement learning approach, in which the action space is a set of cost-improving local moves, could be successful.}

Beyond vehicle routing, we believe our approach can be applied to approximate dynamic programming problems with a combinatorial action space and a neural network approximating a cost-to-go or $Q$ function. Even the simpler case of continuous medium-dimensional action spaces is a challenge. In \cite{ryu2019caql}, a similar approach to ours was competitive with state-of-the-art methods on the DeepMind Control Suite \cite{tassa2018deepmind}.

\section*{Broader Impact}

This paper presents methodological work and does not have foreseeable direct societal implications.

\begin{ack}
We would like to thank Vincent Furnon for his help in selecting OR-Tools parameters to significantly improve baseline performance.
\end{ack}

\bibliographystyle{unsrt}


\newpage

\appendix

\section{Algorithms, hyperparameters, and code}

In this section, we recap key hyperparameters in our approach, as well as their default values.

\begin{enumerate}
    \item Overall parameters:
    \begin{itemize}
        \item Number of policy iterations: we vary this in the paper, but our default value is 250.
        \item Data retention factor: our default value is 1 (data from previous iterations is not decayed in the mean-squared error training objective).
    \end{itemize}
    \item Learning parameters:
    \begin{itemize}
        \item Neural network architecture: we use a fully-connected neural network, with an input layer of size $n$ (binary input), a single hidden layer with 16 ReLU-activated nodes, and a single linear-activated output node.
        \item Learning rate and batch size: we optimize the neural network parameters using standard batch stochastic gradient descent, with a fixed learning rate (default value \num{5e-4}) and batch size (default value 10).
        \item Epochs: we typically pass through the entire dataset 500 times.
        \item LASSO regularization: given all neural network weights as a vector $w$, we penalize the training objective with the LASSO term $\lambda/ \cdot \norm{w}_1 / (\text{number of weights})$, where the default value of $\lambda$ is 0.1. We note that when turning off LASSO regularization (not recommended), we can increase the learning rate to \num{1e-3} and decrease the number of epochs to 300, affording up to almost 2x training speedup. For models with fewer than 16 neurons, we do not use any regularization (i.e. $\lambda = 0$).
    \end{itemize}
    \item Evaluation parameters:
    \begin{itemize}
        \item Number of sample paths to evaluate: we select $N$ random start states to initialize sample path computation. The larger the value of $N$, the more data we obtain to train the cost-to-go estimator. We typically set $N=10$, a very small value since our ablation studies suggest there is more to gain from many policy iterations than from many data points.
        \item Zeroing threshold: when solving the action selection problem, we integrate the neural network modeling the cost-to-go into a MIP formulation. MIP solvers can run into numerical issues in the presence of very small coefficients, so we round every coefficient with absolute value less than a threshold (typically $0.001$) to zero. We have observed that this has no measurable effect on solution quality and greatly reduces the chance of the solver failing.
        \item Time limit: we impose a time limit of 60 seconds on the MIP solver when solving the action selection problem. The solver will then return the best solution found so far (but not necessarily the optimal one). The time limit is rarely reached. On the large instances (the random instances with 51 cities and CVRPLIB instances), we use a time limit of 600 seconds.
        \item Lower bounds: at the evaluation step, we can include simple combinatorial lower bounds. The default setting is to include them. We describe these lower bounds in more detail in a later section of this appendix.
        \item Cuts: a key way to improve MIP runtime and solution quality is the addition of ``cuts'', inequalities that sharpen the solver's ability to prune the search tree using linear programming (LP) bounds. In the default configuration, we add cuts to sharpen the formulation of the neural network argmin problem following the approach detailed in \cite{anderson2020strong}. We also follow a lazy constraint scheme for constraints~\eqref{eq:cutset-in} and \eqref{eq:cutset-out} in which we add violated constraints at every node in the branch-and-bound tree rather than only at integer solutions. Finally, we use a linear programming preprocessing routine to update bounds.
    \end{itemize}
\end{enumerate}

The parameters above describe a setting in which each policy iteration is quick and we continuously improve the policy with a small amount of data. If parallel computing infrastructure is available, we consider a ``high parallelism'' mode, in which we increase the number of sample paths to 200 per iteration over 100 policy iterations.

The experiments on CVRPLIB and the 51 city random instances were run in high parallelism mode, while the ablations on 11 and 21 cities were run with the default settings (and the changes listed in each ablation).

\paragraph{\aedit{Algorithm overview.}}
\aedit{
Algorithm~\ref{algo} presents an overview of our policy iteration scheme in algorithm block form. We note that this is a simplified description of our method, which does not include all refinements, e.g., the mechanism to conserve data from iteration to iteration.

\begin{algorithm}
\caption{High-level overview of our policy iteration scheme. The inputs are simply the problem parameters, namely demands $d$, distances $\Delta$, capacity $Q$, and the number of policy iterations $K$.} \label{algo}
\begin{algorithmic}[1]
\Function{SolveCVRP}{$d, \Delta, Q, K$}
\State $\pi\gets\pi^0$ \Comment{Initialize policy}
\For{$k=1$ to $K$}
\State $\mathcal{D}\gets\emptyset$ \Comment{Initialize empty dataset}
\For{$i=1$ to $N_k$}
\State Select a random state $s_0$
\State $\mathcal{D}\gets\mathcal{D}~\cup$ \Call{EvaluatePolicyFromState}{$\pi$, $s_0$, \texttt{false}} \Comment{Add new data}
\EndFor
\State Use dataset $\mathcal{D}$ to train a new value function approximation and update the policy $\pi$
\EndFor
\State Define $s_0$ as the state where all cities are unvisited
\State \textbf{return} \Call{EvaluatePolicyFromState}{$\pi$, $s_0$, \texttt{true}}
\EndFunction
\Function{EvaluatePolicyFromState}{$\pi$, $s_0$, $x$}
\State $c\gets 0$, $s\gets s_0$
\While{$s$ is not terminal}
\State $a\gets \pi(s)$
\State $c \gets c + C(a)$
\State $s \gets T(s, a)$
\EndWhile
\If{$x = \texttt{true}$}
\State \textbf{return} the list of selected actions (routes) and the total incurred cost
\Else
\State \textbf{return} all visited states and the total cost incurred from each one, denoted $\{s^i, c^i\}_i$
\EndIf
\EndFunction
\end{algorithmic}
\end{algorithm}

Our code is available as part of the \texttt{tf.opt} repository: \url{https://github.com/google-research/tf-opt}.
}

\paragraph{\aedit{Optimizing over a trained neural network.}}
\aedit{The second term of the objective~\eqref{eq:objective} is a nonlinear function (fully-connected one-layer neural network with ReLU activations), and nontrivial to model using mixed-integer linear programming. We use a technique developed by Anderson et al. \cite{anderson2020strong} to model $\hat{V}(t)$. For clarity, assume that $\hat{V}(\cdot)$ has a single hidden layer, with $P$ hidden nodes, each with a ReLU activation.

Let $w^{(p)}\in\mathbb{R}^n$ designate the vector of weights, and $b^{(p)}\in\mathbb{R}$ the bias term, for the $p$-th hidden node. Define $w^{\text{output}}\in\mathbb{R}^P$ and $b^{\text{output}}\in\mathbb{R}$ analogously for the output layer. For any vector of weights $w$, let $\text{supp}(w)$ indicate the set of indices $i$ such that $w_i\neq 0$. Finally, define $\rho:\mathbb{R}\to\{0,1\}$ a modified version of the sign function, where $\rho(a)$ returns 1 if $a\ge 0$, and 0 otherwise. Then we can write the problem $\min_{t\in\{0,1\}^n}\hat{V}(t)$ as
\begin{subequations}
\begin{align}
    \min\quad & \hat{V}(t) &\coloneqq& \sum_{p=0}^{P-1} w^{\text{output}}_p y^{(p)}+b^{\text{output}}\\
    \text{s.t.}\quad & y^{(p)} &\ge&~w^{(p)}\cdot t + b^{(p)} & \forall 0 \le p < P \\
    & y^{(p)} &\le&~w^{(p)}\cdot t + b^{(p)} - M_{-}^{(p)}(1-z^{(p)}) & \forall 0 \le p < P \label{eq:Mminus}\\
    & y^{(p)} &\le&~M_{+}^{(p)}z^{(p)} & \forall 0 \le p < P \label{eq:Mplus}\\
    & y^{(p)} &\le& \sum_{i\in I}w_i^{(p)}\left(t_i - \left(1 - \rho\left(w_i^{(p)}\right)\right) \left(1-z^{(p)}\right)\right) +\\
    &&& \left(b^{(p)}+\sum_{i\notin I}w_i^{(p)}\rho\left(w_i^{(p)}\right)\right)z^{(p)} & \forall 0\le p < P, I\subseteq \supp w^{(p)} \label{eq:exponential}\\
    & t &\in&~\mathcal{T}\\
    & y^{(p)} &\in& ~\mathbb{R}&\forall 0 \le p < P\\
    & z^{(p)} &\in& ~\{0, 1\}&\forall 0 \le p < P.
\end{align}
\end{subequations}

Notice that in addition to the $n$ initial binary decision variables $t_i$, we define two additional decision variables for each hidden node in the neural network. One is continuous and models the output of the hidden node, the other is binary and indicates whether the pre-activation function is positive or negative (i.e. whether the ReLU is active or not). This relationship is enforced by the ``big-$M$'' constraints~\eqref{eq:Mminus} and \eqref{eq:Mplus}. In principle, any large enough values of $M_+$ and $M_-$ can enforce this relationship, but values that are too large weaken the formulation and can decrease tractability. We therefore compute these big-$M$ values for each hidden node by maximizing (or minimizing, depending on whether computing $M_-$ or $M_+$) the pre-activation function over the LP relaxation of the neuron's input domain (which includes all the problem specifications, including the PC-TSP and knapsack constraints). This operation can be performed iteratively, neuron-by-neuron, to compute suitable values for $M_+^{(p)}$ and $M_-^{(p)}$.

The set $\mathcal{T}$ encapsulates any other constraints on the binary variables $t$ (input domain of the neural network $\hat{V}(\cdot)$), in particular the ones associated with the prize-collecting traveling salesman formulation~\eqref{eq:formulation}.

We note that this formulation is not polynomial in size, as there are exponentially many constraints of type~\eqref{eq:exponential}. However, these constraints are not needed for correctness, they simply strengthen the formulation. When solving the action selection problem, we generate these constraints on the fly, using a linear-time separation oracle as described in~\cite{anderson2020strong}.
}

\paragraph{\aedit{Local search warm start.}}
\aedit{In order to speed up the MIP solve in the action selection problem~\eqref{eq:formulation}, we provide the solver with an initial primal-feasible solution, obtained via a simple local search heuristic. This warm start ensures we always obtain a feasible solution and allows the solver to spend more time in the branch-and-bound tree. As a result, we can set a lower solver time limit, and increase the number of policy iterations performed in a fixed amount of time.

Our local search heuristic (1-OPT with random start) can be described as follows. We first create a random feasible route, by randomly sampling unvisited cities until we hit the capacity constraint. We then consider every unvisited city, and compute the cost of removing this city in the route (if it is already included) or including it in the route (if it is not). For each of the $\mathcal{O}(n)$ unvisited cities, evaluating the cost requires one call to a TSP solver (to evaluate the route distance) and one neural network evaluation (to evaluate the future cost of unvisited cities). Both are computationally tractable, and allow us to quickly perform many iterations, possibly with several random restarts.}

\paragraph{Combinatorial lower bounds.} In the main text, we mention refining the cost-to-go $\hat{V}(\cdot)$ with combinatorial lower bounds linear in the decision variables of problem~\eqref{eq:formulation}. We now describe each bound in more detail -- note that they each assume the distances respect the triangle inequality (metric vehicle routing):
\begin{enumerate}
    \item Maximum out-and-back bound: given a state $s$, the cost-to-go cannot be exceeded by the distance to and from the furthest city from the depot, i.e.,
    \[
    V(s)\ge \max_{i:s_i=1}(\Delta_{0i}+\Delta_{i0}).
    \]
    This bound is rather weak for a large number of remaining cities, but it is tight when a single city remains.
    \item Shortest-edges bound: given a state $s$, we must take at least one edge into and out of each city, thus paying at least half the cost of the shortest edges into and out of each city, i.e.,
    \[
    V(s) \ge \sum_{i:s_i=1}\min_{j:s_j=1;j\neq i}\frac{1}{2}(\Delta_{ji} + \Delta_{ij}).
    \]
    \item Refined shortest-edges bound: we can refine the bound above using demand information to incorporate a bound on the minimum number of vehicles required (and thus the minimum number of times the depot must be visited), i.e.,
    \[
    V(s) \ge \sum_{i:s_i=1,i>0}\min_{j:s_j=1;j\neq i}\frac{1}{2}(\Delta_{ji} + \Delta_{ij}) + \frac{1}{2}\left\lceil\frac{\sum_{i:s_i=1,i>0}d_i}{Q}\right\rceil\min_{j:s_j=1;j>0}\Delta_{0j}.
    \]
\end{enumerate}

\section{CVRP instances}

In this section, we describe the CVRP instances we use to evaluate our reinforcement learning framework.

\paragraph{Random Euclidean instances.} We follow the generation procedure of Nazari et al.\ (2018) to construct random instances. We sample $n$ locations uniformly at random in the unit square, then define the distance $\Delta_{ij}$ to be the Euclidean distance from city $i$ to city $j$ (symmetric). One of the cities is randomly selected to be the depot, and the demand for the remaining $n-1$ cities is uniformly sampled from $\{1,2,\ldots,9\}$. The vehicle capacity $Q$ scales with the number of cities, with $Q=20$ for $n=11$, $Q=30$ for $n=21$, and $Q=40$ for $n=51$.

\paragraph{Standard library instances.} An issue with evaluating algorithms uniform Euclidean vehicle routing instances is that such instances are known to satisfy certain strong regularity properties that may not be manifested in real-world problems \cite{beardwood1959shortest,goemans1991probabilistic}. In an effort to benchmark methods against more realistic instances, the CVRP library (CVRPLIB) is an online resource cataloging instances from the literature---either real problems, or synthetic examples inspired by certain properties of real-world instances. We include 50 instances from CVRPLIB (``A'' and ``B'' instances, ranging in size from 32 to 78 cities) \cite{augerat1995computational}. We note that in many cases, the optimal values for these problems are known, and listed on the CVRPLIB website \cite{uchoa2017new}. However, we do not use these optimal values because they consider a slightly different setting in which the number of vehicles is fixed, and thus overestimate the true optimum in our unbounded-fleet setting.

\section{Baselines}

In this section, we briefly describe the baselines against which we compare our results.

\paragraph{OR-Tools.} Google's open-source OR-Tools library is an often-used benchmark for combinatorial optimization problems, and incorporates one of the best existing vehicle routing solvers \cite{or-tools}, which combines exact approaches with local search and other heuristics to provide high-quality solutions. On instances of moderate size such as the ones presented in this paper, OR-Tools configured to perform a few minutes of local search typically produces optimal or near-optimal solutions, \aedit{motivating our use of OR-Tools as a reference point when an optimal approach does not scale}. The results we report with OR-Tools are considerably better than those reported in previous reinforcement learning for vehicle routing papers \cite{nazari2018reinforcement,Kool2019}. This reflects the fact that we configure OR-Tools for solution quality and not for speed, in contrast to other methods.

\paragraph{\aedit{Optimal.}} \aedit{Vehicle routing problems can be solved to optimality up to a certain problem size using mixed-integer programming (MIP) or constraint programming (CP) approaches. In addition to a routing solver, the OR-Tools library also provides a constraint programming solver called CP-SAT. To avoid confusion and remain consistent with the literature on reinforcement learning for vehicle routing, we refer to the routing solver as ``OR-Tools'' and the CP-SAT solver as ``CP-SAT'', even though both are technically part of OR-Tools.

We use CP-SAT to compute both feasible solutions and lower bounds. On small instances, the lower bounds coincide with the best feasible solution, yielding a certificate of optimality. When tractable, these provably optimal solutions are a reference point for our results, and we use the gap to the optimal solution as a key metric of success. Unfortunately, CP-SAT cannot prove optimality in a reasonable time for $n=51$ cities. For larger instances, we therefore fall back to the near-optimal OR-Tools (routing) solver as a reference point. We note that as CP-SAT and the OR-Tools routing solvers only work on integer distances, we multiply all problem distances by $10^4$ and round to the nearest integer, so results are accurate within $10^{-4}$.}

\paragraph{Greedy.} One of the claims of this paper is that our method performs well on CVRP instance both because of the combinatorial structure of the action space and because of the machine learning model's ability to \aedit{learn} the complexity of the value function. A useful benchmark for our approach is therefore a method which features only the combinatorial action space without the learning component. This approach involves solving problem~\eqref{eq:formulation}, replacing the cost-to-go $\hat{V}(t)$ with the following trivial upper bound:
\[
UB(t)=\sum_{i=0}^{n-1}(\Delta_{0i} + \Delta_{i0})t_i,
\]
representing the total distance of covering all remaining cities with one route per city. We refer to this baseline as a greedy method, because it overestimates the cost-to-go and thus compels the optimal action to pack as many cities as possible (and particularly cities far from the depot).

\paragraph{Existing approaches.} Finally, we compare our approach with existing RL approaches for vehicle routing \cite{nazari2018reinforcement,Kool2019}. This comparison is less direct than the two above, because both approaches consider a multi-instance learning setting, where the model is trained on a large sample of problem instances and evaluated on unseen instances from the same distribution. As a result, both papers \cite{nazari2018reinforcement,Kool2019} report out-of-sample solution quality metrics, whereas we train and evaluate on one instance at a time. Both frameworks are valuable: on one hand, learning insights from multiple instances that can be generalized to a new problem can significantly improve solve times; on the other hand, real-world vehicle routing problems do not typically come with distributional information, rendering prior learning irrelevant. Though they are not directly comparable, we nevertheless display solution quality metrics for both of these methods along with ours, as evidence that our framework produces solutions in the same ballpark as other RL frameworks.

\section{Additional simulation results}

\begin{figure}
    \centering
    \begin{subfigure}{0.45\columnwidth}
        \includegraphics[width=\columnwidth]{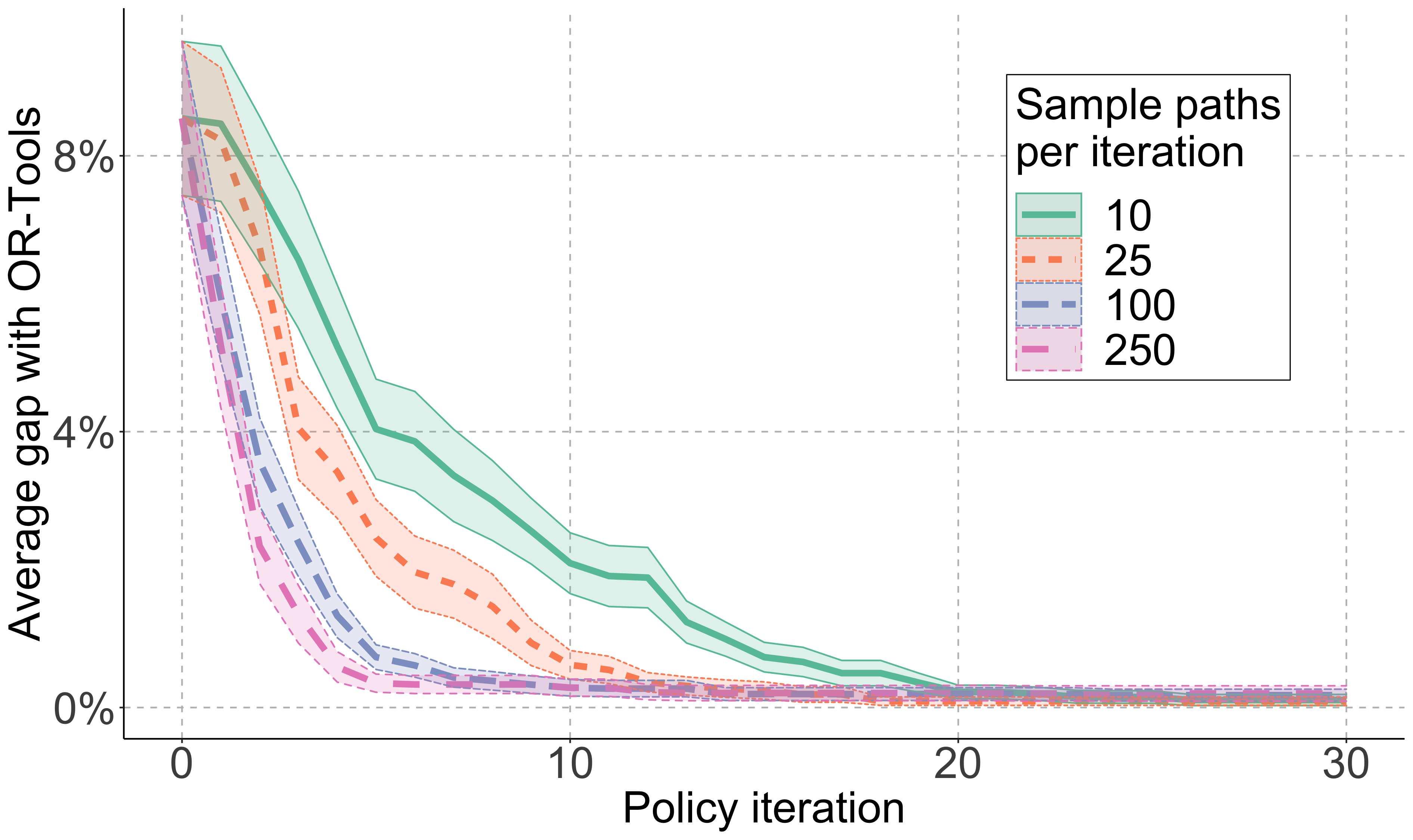}
        \caption{$n = 11$}
        \label{fig:parallel_11}
    \end{subfigure}\hfill
    \begin{subfigure}{0.45\columnwidth}
        \includegraphics[width=\columnwidth]{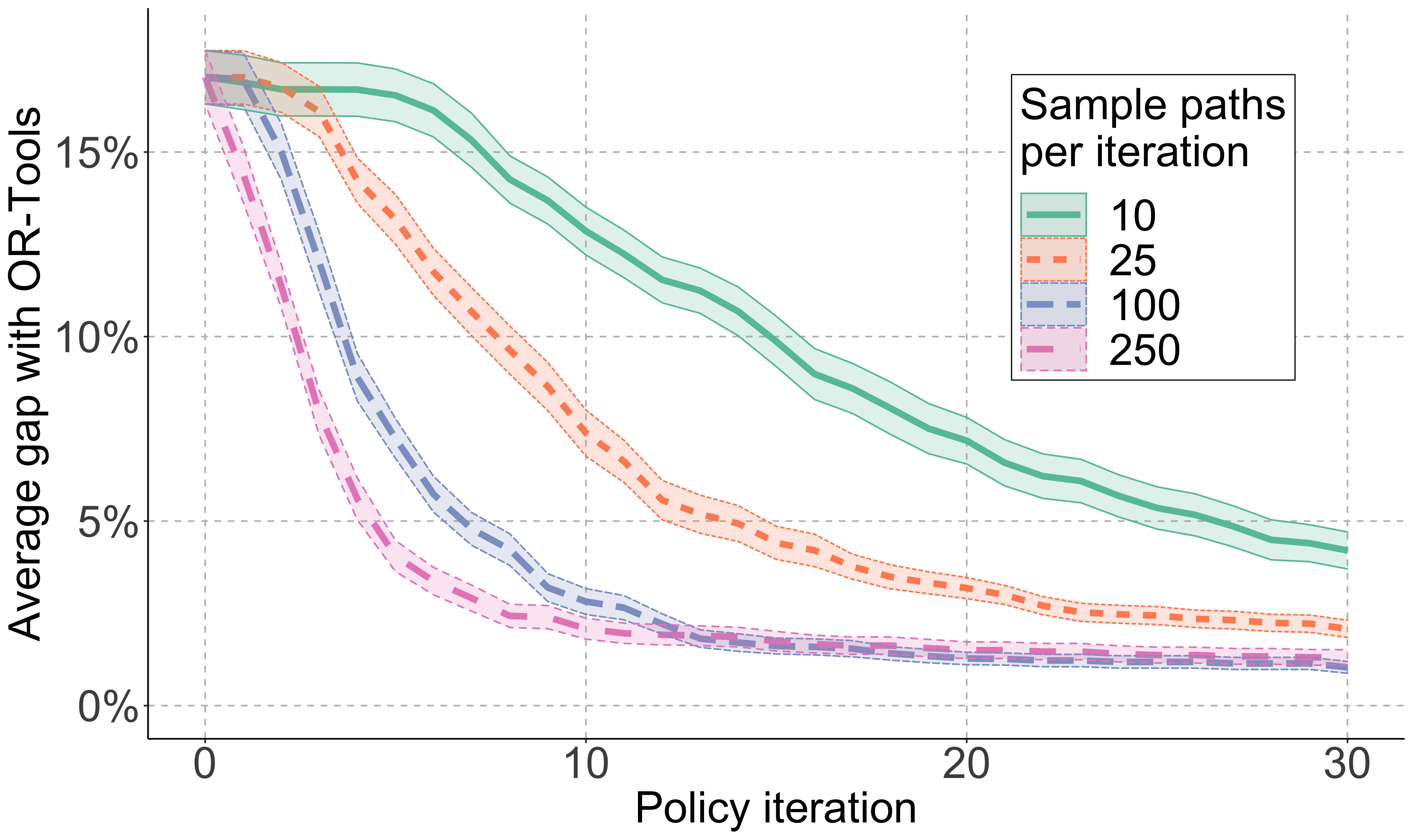}
        \caption{$n = 21$}
        \label{fig:parallel_21}
    \end{subfigure}\hfill
    \caption{Effect of number of sample paths per iteration on the gap at each iteration ($\gamma=1$). Results averaged over 50 random Euclidean instances, error bars show standard errors (SEM).}
    \label{fig:parallel}
\end{figure}

In Section A of the appendix, we included a description of our hyperparameters, and in particular identified a ``high parallelism'' setting. We now provide some experimental justification for this setting in Figure~\ref{fig:parallel}. We see that increasing the number of sample paths per iteration can provide significant improvements on a per-iteration basis. Computing sample paths is a highly parallelizable task, so depending on the computing platform available, it may be more efficient to consider a small number of policy iterations, and compensate by computing many sample paths in parallel at each iteration.

\end{document}